\documentclass[10pt,twocolumn,letterpaper]{article}

\usepackage{cvpr}              

\usepackage{graphicx}
\usepackage{float}

\definecolor{cvprblue}{rgb}{0.21,0.49,0.74}
\usepackage[pagebackref,breaklinks,colorlinks,allcolors=cvprblue]{hyperref}
\usepackage{algorithm}
\usepackage{algpseudocode}
\usepackage{multirow}

\def\paperID{*****} 
\def\confName{CVPR}
\def\confYear{2025}

\title{The Multimodal Paradox: How Added and Missing Modalities Shape Bias and Performance in Multimodal AI}

\author{
Kishore Sampath\\
Northeastern University\\
Boston, MA, USA\\
{\tt\small sampath.ki@northeastern.edu}
\and
Pratheesh\\
Northeastern University\\
Boston, MA, USA\\
{\tt\small lnu.prat@northeastern.edu}
\and
Ayaazuddin Mohammad\\
Northeastern University\\
Boston, MA, USA\\
{\tt\small mohammad.ay@northeastern.edu}
\and
Resmi Ramachandranpillai \\
Institute for Experiential AI \\ 
Northeastern University\\
Boston, MA, USA\\
{\tt\small r.ramachandranpillai@northeastern.edu}
}

\begin{document}
\maketitle
\begin{abstract}
Multimodal learning, which integrates diverse data sources such as images, text, and structured data, has proven to be superior to unimodal counterparts in high-stakes decision-making. However, while performance gains remain the gold standard for evaluating multimodal systems, concerns around bias and robustness are frequently overlooked. In this context, this paper explores two key research questions (RQs): (i) RQ1 examines whether adding a modality consistently enhances performance and investigates its role in shaping fairness measures, assessing whether it mitigates or amplifies bias in multimodal models; (ii) RQ2 investigates the impact of missing modalities at inference time, analyzing how multimodal models generalize in terms of both performance and fairness. Our analysis reveals that incorporating new modalities during training consistently enhances the performance of multimodal models, while fairness trends exhibit variability across different evaluation measures and datasets. Additionally, the absence of modalities at inference degrades performance and fairness, raising concerns about its robustness in real-world deployment.
We conduct extensive experiments using multimodal healthcare datasets containing images, time series, and structured information to validate our findings. 
\end{abstract} 
\vspace{-0.5cm}
\section{Introduction}
\label{sec:intro}
The world is inherently multimodal \cite{baltruvsaitis2018multimodal}, with human experiences shaped by diverse streams of information—ranging from visual and textual cues to auditory signals and structured data. These modalities do not exist in isolation but interact in complex ways, influencing perception, decision-making, and learning. 

Multimodal learning leverages the inherent interplay of diverse modality sources to make more informed and impactful decisions, particularly in high-stakes applications such as healthcare \cite{kline2022multimodal}. However, current research in multimodal AI often focuses on performance enhancement while neglecting thorough evaluations of bias \cite{Alpher01, ramachandranpillai2024fairness} and generalization, which are essential for ensuring models remain reliable and robust in deployment.

Limited research has been conducted on evaluating biases in multimodal data and models, particularly in applications such as interview assessments \cite{10.1145/3382507.3418889, Booth} and stages like multimodal fusion \cite{lu2024collaborative}. In medical domains, including image analysis and clinical decision-making in dermatology \cite{lopez2025generative}, the primary focus has been on addressing the underrepresentation of patient populations. However, these studies largely concentrate on mitigating bias at the data level or during specific model stages, often overlooking the broader picture of how fluctuations in modality availability—across both training and inference—affect model performance and fairness. Specifically, we explore how variations in available modalities influence the model's performance, fairness, and robustness, addressing an important gap in the literature. 

\textbf{Contributions}: Aiming to understand the performance, fairness, and robustness of multimodal models in real-time applications, we formulate two key research questions (RQs):
\begin{enumerate}
    \item \textit{RQ1: How does the compounding effect of introducing new
modalities in training impact both the performance and bias of the 
model in downstream decision-making?}
\item \textit{RQ2: How do missing modalities at inference impact performance and fairness of the model at deployment?}
\end{enumerate}
To ground our findings, we leverage the PhysioNet \footnote{https://physionet.org/} multimodal datasets, MIMIC-Eye \cite{Hsieh2023} and MIMIC-IV-Ext-MDS-ED \cite{LopezAlcaraz2024} which include a combination of images, text, and structured information, providing a rich and complex source of data. Studies involving these intricate multimodal datasets are limited in the literature due to the challenging relationships between the modalities, which require sophisticated methods to process and interpret effectively. 

\begin{figure*}[t!]
    \centering
    \includegraphics[width=0.7\textwidth,height=6cm]{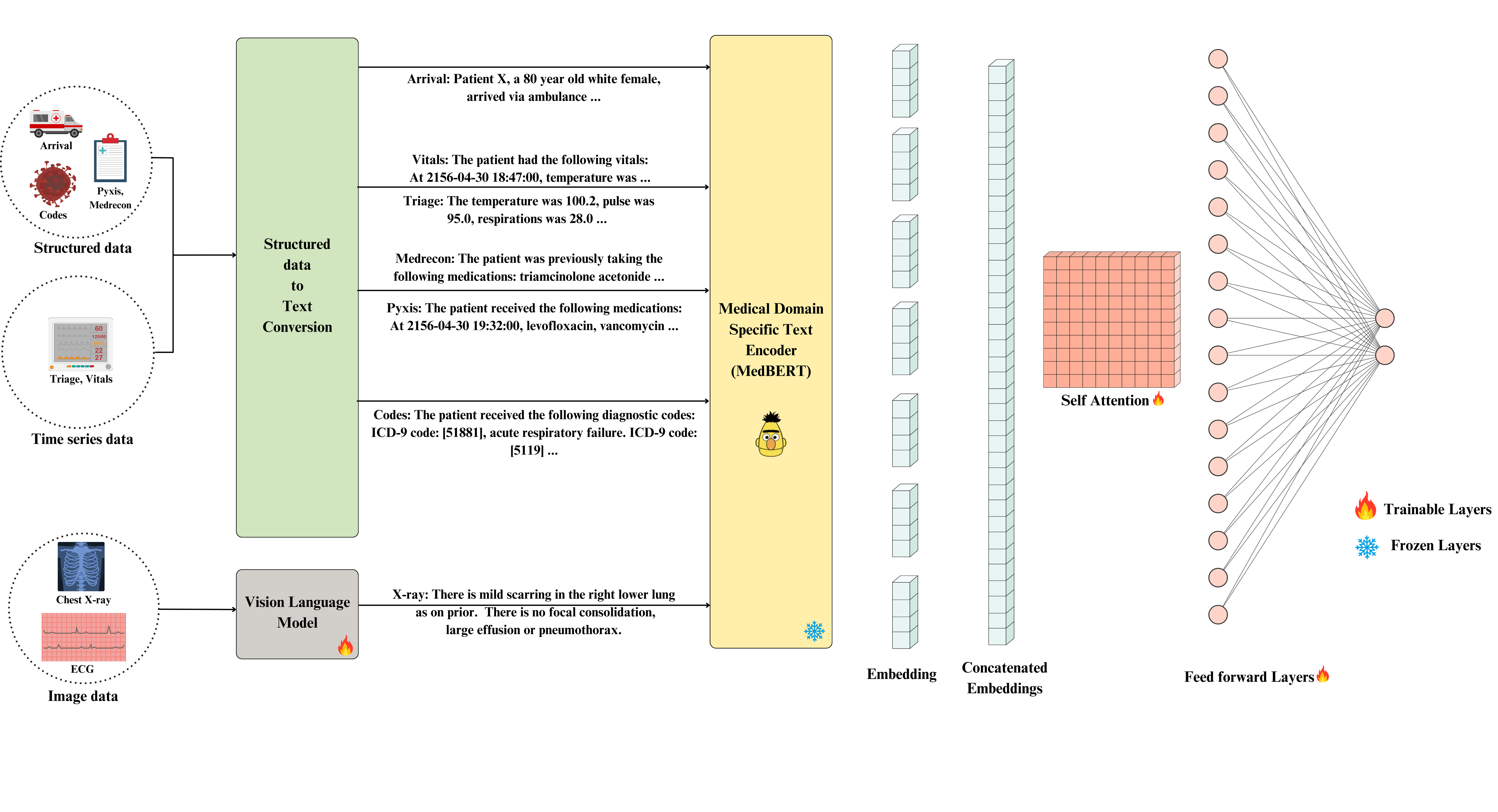}
    \vspace{-0.8cm} 
    \caption{A multimodal framework for unified representation learning transforming heterogeneous clinical inputs (structured, time series and imaging data) into a common text-based feature space for downstream diagnostic classification tasks.}
    \label{fig:arch}
    \vspace{-0.2cm}
\end{figure*}



\label{sec:formatting}

\section{RQ1: How does the compounding effect of introducing new modalities in training impact both the performance and bias of the model in downstream decision making?}

\vspace{-5pt}

A fundamental assumption in multimodal learning is modality monotonicity - the idea that adding an additional modality consistently improves performance \cite{he2024efficient}. Although this holds in many cases, the effect of added modalities on fairness is less understood. Introducing a new modality can potentially enhance predictive accuracy; however, it may also introduce spurious correlations or exacerbate existing biases. Conversely, the addition of a modality may have no significant impact on performance but could improve the fairness of the model. Therefore, the central question arises: Does the inclusion of a specific modality justify the resulting performance/fairness gains, or does it come at the expense of fairness?

To address this question, we propose a structured evaluation approach. First, we analyze the performance gains for each individual modality to understand its specific contribution to the overall model. Next, we incrementally add modalities during the training process to assess how the inclusion of each new modality impacts the model's predictions. By comparing these models, we can analyze whether performance gains follow a monotonic trend and whether the introduction of additional modalities disproportionately skews predictions. This method allows us to examine both the individual and combined effects of modality additions, providing deeper insights into how each modality contributes to the model's overall behavior.



\section{RQ2: How do missing modalities at inference impact performance and fairness of the model at deployment?}
In RQ2, we shift our focus to model robustness when one or more modalities are unavailable at inference time. In real-world applications, missing data is a common problem, particularly in healthcare, where some patients may lack complete diagnostic reports due to financial limitations, incomplete medical records, or restrictions in accessing specialized tests. Our study will measure how different multimodal architectures adapt to such data gaps, analyzing their robustness to performance stability and fairness when trained modalities are absent. To assess this, we randomly mask a subset of the modalities and evaluate both model performance and bias. The pseudocode for our masking procedure is provided in Algorithm 1.

\begin{algorithm}[]
\footnotesize
\caption{\small{Random Modality Masking}}
\label{alg:modality_masking}
\begin{algorithmic}[1]
\State \textbf{Input:} Modalities $\{\mathbf{x}_1, \mathbf{x}_2, \dots, \mathbf{x}_M\}$ \Comment{$M$ is the number of modalities}
\State \textbf{Parameters:} Probability $p$ for all modalities
\State \textbf{Output:} Masked modalities $\{\mathbf{x}_1', \mathbf{x}_2', \dots, \mathbf{x}_M'\}$

\For{$i = 1$ to $M$}
    \State Sample $b_i \sim \text{Bernoulli}(p)$ \Comment{Sample a binary mask for modality $i$}
    \State $\mathbf{x}_i' \gets b_i \cdot \mathbf{x}_i$ \Comment{Apply masking: retain if $b_i = 1$, mask if $b_i = 0$}
\EndFor

\State \textbf{Return:} $\{\mathbf{x}_1', \mathbf{x}_2', \dots, \mathbf{x}_M'\}$
\end{algorithmic}
\end{algorithm}

For each data instance, we independently sample a binary mask for every modality using a Bernoulli distribution with probability $p$. The model then generates predictions based on the masked input, and we evaluate both its performance and fairness under these conditions.



\section{Modality Integration and Prediction}

In this section, we present our multimodal framework, which integrates information from heterogeneous modalities including images, ECGs, time series, and structured data into a unified feature space. The extracted features are then fused for downstream classification tasks.

Let $D = \{X_i, Y_i\}$ for $i = 1$ to $n$ denote our dataset, where $X_i = \{x_{i1}, x_{i2}, x_{i3}, ..., x_{im}\}$ represents the $m$ different modalities for the $i$-th patient, and $Y_i$ is the corresponding label. Here, $n$ is the total number of patients in our multimodal dataset. Our framework employs a consistent architectural approach: the processed information from each modality is embedded into a unified feature space using a domain-specific medical text encoder (MedBERT) \cite{Rasmy2021}. The resulting modality-specific embeddings are concatenated and processed through a self-attention module to capture cross-modal interactions, followed by a feedforward network and a classification head for final predictions. The overall architecture is illustrated in Figure 1. In the following, we describe how each modality has been represented within the unified feature space. 

\textbf{Structured information.}
To convert patient medical history, vital signs, and laboratory results into textual format we follow \cite{lee2024emergency}, which utilizes a conventional serialization approach similar to SmartPhrases/DotPhrases in the Epic EHR system \cite{chang2021emr}, facilitating the precise insertion of relevant clinical information within a structured text format.

\textbf{Chest X-rays.}
For transcribing chest X-rays into radiology reports, we fine-tune the Qwen 2.5 3B Vision-Language (VL) model \cite{qwen2025qwen25technicalreport} using QLoRA with 4-bit integer quantization. Specifically, we initialized the Qwen 2.5 3B VL model from pre-trained checkpoints and applied 4-bit normalized float (NF4) quantization for parameter-efficient fine-tuning. The model was fine-tuned for three epochs using Low-Rank Adaptation (LoRA) with a rank of 8 and a scaling factor of 16. Dropout rate of 0.05 was applied to the query and value projection layers. The training utilized 3689 chest X-ray images from the MIMIC-EYE dataset, partitioned into a 75\%/25\% training-validation split. Optimization was performed using AdamW with a learning rate of 2e-4 and a warm-up ratio of 0.03. Training was carried out with a batch-size of 8 and a gradient accumulation of 4 steps, requiring approximately 3.7 hours of computation on a single NVIDIA A100 GPU. This model is chosen for its top ranking on the OpenVLM leaderboard among vision-language models (VLMs) with under 4 billion parameters.. 

\textbf{Electrocardiogram (ECG)}
To generate textual descriptions of ECG signals from MIMIC-IV-ECG, we employ PULSE-7B, a multimodal large language model (MLLM) \cite{liu2024teachmultimodalllmscomprehend} specifically designed for ECG image interpretation.

\section{Experiments}
\subsection{Settings}
\paragraph{Downstream tasks.} For the MIMIC-EYE dataset, we consider a binary classification task to predict emergency department (ED) disposition, where 1 indicates hospital admission and 0 signifies discharge to home. For the MIMIC-IV-ECG dataset, we selected a random subset of data to predict whether a patient requires intensive care unit (ICU) admission within the next 24 hours.

\textbf{Fairness measures.} Following \cite{chen2024unmasking} in the healthcare domain, we consider Demographic Parity (DP) and True Positive Rate(TPR) as fairness metrics in our experiments.

\subsection{Results}

\subsubsection{RQ1}
\begin{table}[t]
\centering
\caption{MIMIC-EYE modality-wise performance and fairness analysis}
\scriptsize
\setlength{\tabcolsep}{1.5pt}
\renewcommand{\arraystretch}{1.1}
\begin{tabular}{@{}lccccccc@{}}
\hline
& \multicolumn{7}{c}{\textbf{MIMIC-EYE}} \\
\textbf{Category} & \textbf{Arrival} & \textbf{Triage} & \textbf{Medrecon} & \textbf{Vitals} & \textbf{Pyxis} & \textbf{Codes} & \textbf{X-ray} \\
\hline
\multicolumn{8}{l}{\textbf{Performance}} \\
F1 Score & \textbf{0.920} & 0.742 & 0.745 & 0.693 & 0.766 & 0.830 & 0.756 \\
AUC-ROC & \textbf{0.908} & 0.764 & 0.640 & 0.722 & 0.798 & 0.854 & 0.513 \\
AUPRC & 0.887 & 0.823 & 0.719 & 0.802 & 0.866 & \textbf{0.893} & 0.644 \\
Balanced Accuracy & \textbf{0.868} & 0.688 & 0.558 & 0.658 & 0.723 & 0.753 & 0.499 \\
\hline
\multicolumn{8}{l}{\textbf{DP (race)}} \\
White & 0.700 & 0.586 & 0.849 & 0.517 & 0.558 & 0.678 & \textbf{1.000} \\
Non-White & 0.710 & 0.545 & 0.860 & 0.510 & 0.558 & 0.689 & \textbf{1.000} \\
\hline
\multicolumn{8}{l}{\textbf{DP (gender)}} \\
Female & 0.712 & 0.565 & 0.850 & 0.515 & 0.575 & 0.687 & \textbf{0.999} \\
Male & 0.701 & 0.585 & 0.849 & 0.522 & 0.557 & 0.682 & \textbf{1.000} \\
\hline
\multicolumn{8}{l}{\textbf{TPR (race)}} \\
White & 0.998 & 0.737 & 0.895 & 0.640 & 0.745 & 0.885 & \textbf{0.999} \\
Non-White & 0.999 & 0.691 & 0.909 & 0.630 & 0.724 & 0.882 & \textbf{1.000} \\
\hline
\multicolumn{8}{l}{\textbf{TPR (gender)}} \\
Female & 0.996 & 0.718 & 0.891 & 0.641 & 0.753 & 0.891 & \textbf{0.998} \\
Male & 0.999 & 0.730 & 0.901 & 0.647 & 0.730 & 0.879 & \textbf{1.000} \\
\hline
\end{tabular}
\end{table}

\begin{table}[t]
\centering
\caption{MIMIC-IV-Ext-MDS-ED modality-wise performance and fairness analysis}
\scriptsize
\setlength{\tabcolsep}{1.5pt}
\renewcommand{\arraystretch}{1.1}
\begin{tabular}{@{}lccc@{}}
\hline
& \multicolumn{3}{c}{\textbf{MIMIC-IV-Ext-MDS-ED}} \\
\textbf{Category} & \textbf{ECG} & \textbf{Events} & \textbf{Vitals} \\
\hline
\multicolumn{4}{l}{\textbf{Performance}} \\
F1 Score & 0.174 & 0.272 & \textbf{0.685} \\
AUC-ROC & 0.618 & 0.614 & \textbf{0.799} \\
AUPRC & 0.527 & 0.510 & \textbf{0.740} \\
Balanced Accuracy & 0.530 & 0.538 & \textbf{0.727} \\
\hline
\multicolumn{4}{l}{\textbf{DP (race)}} \\
White & 0.063 & 0.138 & \textbf{0.450} \\
Non-White & 0.047 & 0.104 & \textbf{0.297} \\
\hline
\multicolumn{4}{l}{\textbf{DP (gender)}} \\
Female & 0.064 & 0.144 & \textbf{0.412} \\
Male & 0.067 & 0.135 & \textbf{0.435} \\
\hline
\multicolumn{4}{l}{\textbf{TPR (race)}} \\
White & 0.103 & 0.165 & \textbf{0.694} \\
Non-White & 0.081 & 0.161 & \textbf{0.565} \\
\hline
\multicolumn{4}{l}{\textbf{TPR (gender)}} \\
Female & 0.094 & 0.188 & \textbf{0.685} \\
Male & 0.107 & 0.180 & \textbf{0.684} \\
\hline
\end{tabular}
\end{table}

\textbf{Modality-wise analysis. }First, we perform a modality-wise analysis of performance and bias across both the datasets to evaluate the individual contributions of each modality, as presented in Table 1 and Table 2.

In MIMIC-EYE, Arrival achieves the highest F1 Score (0.920), while Vitals has the lowest (0.693), indicating substantial variability across modalities. Arrival performs best in terms of AUC-ROC (0.854), whereas X-ray struggles with the lowest AUC-ROC (0.513). Similarly, Medrecon and X-ray show lower balanced accuracy, suggesting challenges in their classification. In MIMIC-IV-Ext-MDS-ED, Vitals emerges as the strongest modality across all metrics, with an F1 Score of 0.685 and AUC-ROC of 0.799, while Events lags with an F1 Score of 0.272 and relatively lower AUPRC and balanced accuracy.

Fairness analysis based on DP and TPR indicates minimal disparities but reveals some trends. For race, White and Non-White groups show similar DP values, but Non-White individuals have slightly lower TPR for ECG (0.081 vs. 0.103) and Vitals (0.565 vs. 0.694). For gender, females and males exhibit comparable DP values, though females have a marginally lower TPR for ECG (0.094 vs. 0.107) and Vitals (0.685 vs. 0.684). This analysis evaluates modality contributions and their impact on multimodal performance and fairness.

\begin{figure}[htbp]
    \centering
    \includegraphics[width=0.4\textwidth,height=5cm]{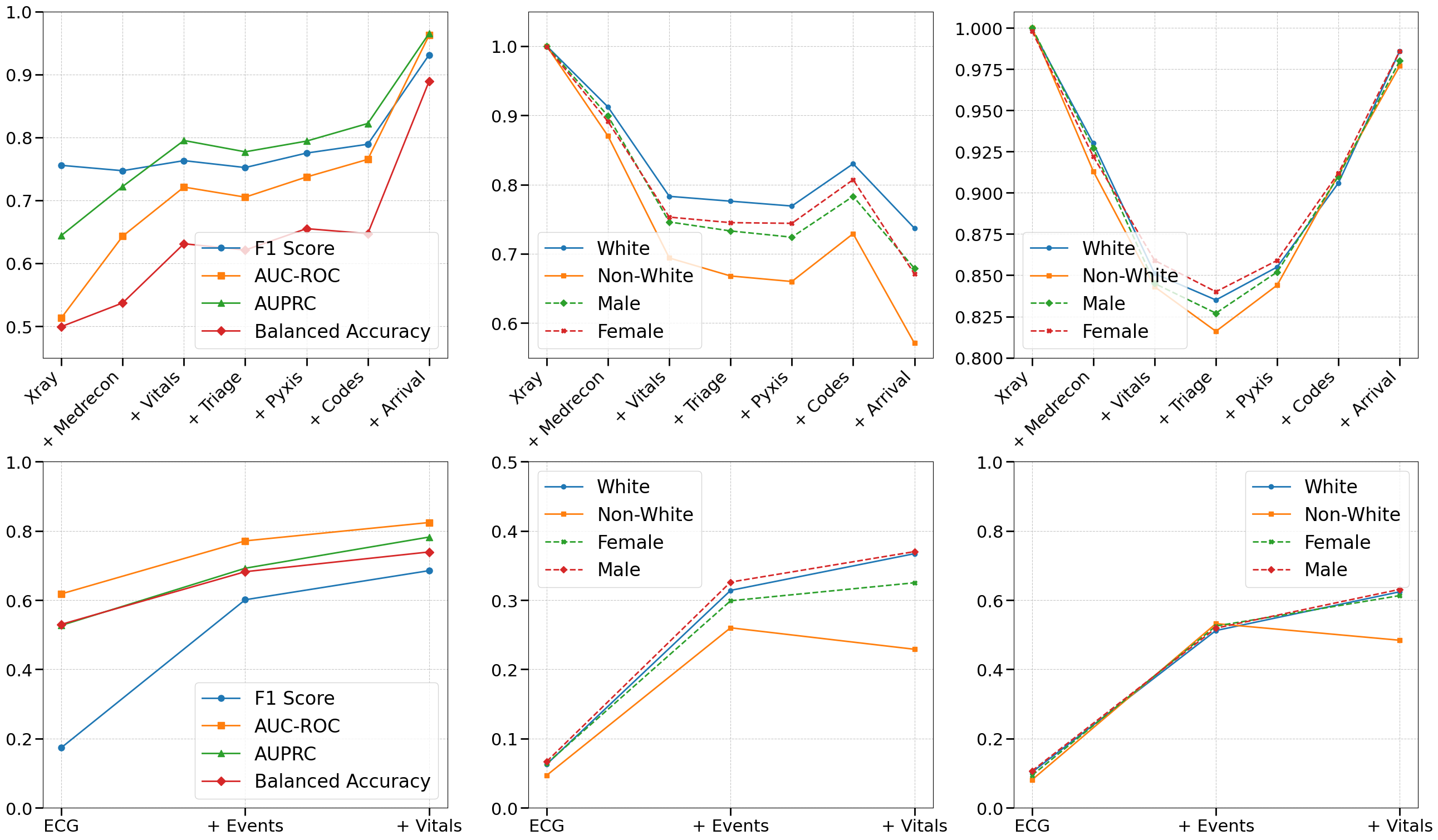} 
    \caption{Impact of incrementally adding modalities during training on performance (left), DP (middle), and TPR (right). First row represents results for MIMIC-Eye and second row for MIMIC-IV-Ext-MDS-ED. X-axis represents Modalities sequentially added to the base ECG model and  Y-axis represents metrics. }
    \label{fig:rq1}
\end{figure}

\textbf{Impact of modality addition during training.} We show in Figure \ref{fig:rq1} the impact of modality addition during training.  The results demonstrate that incorporating additional modalities significantly enhances overall model performance, as reflected in the steady increase in F1 Score, AUC-ROC, AUPRC, and Balanced Accuracy across both the datasets. Initially, relying solely on X-ray/ECG data leads to lower performance, but adding more features progressively improves predictive capability, with the highest performance observed when all modalities are included.

However, fairness disparities fluctuate among the datasets. For MIMIC-Eye, disparity emerges, particularly between White and Non-White groups, with Non-White individuals consistently experiencing lower fairness. In terms of DP, we observe a decreasing trend, with a widening gap between White and Non-White groups, indicating growing disparities as more modalities are added. Meanwhile, for TPR, fairness initially drops but improves with codes and arrival, highlighting their role in mitigating declines. 

In MIMIC-IV-Ext-MDS-ED, we observe a consistent upward trend across DP and TPR, indicating more modalities contribute to both performance and fairness.



\subsubsection{RQ2}
\begin{figure}[hbt!]
    \centering
    \includegraphics[width=0.4\textwidth,height=5cm]{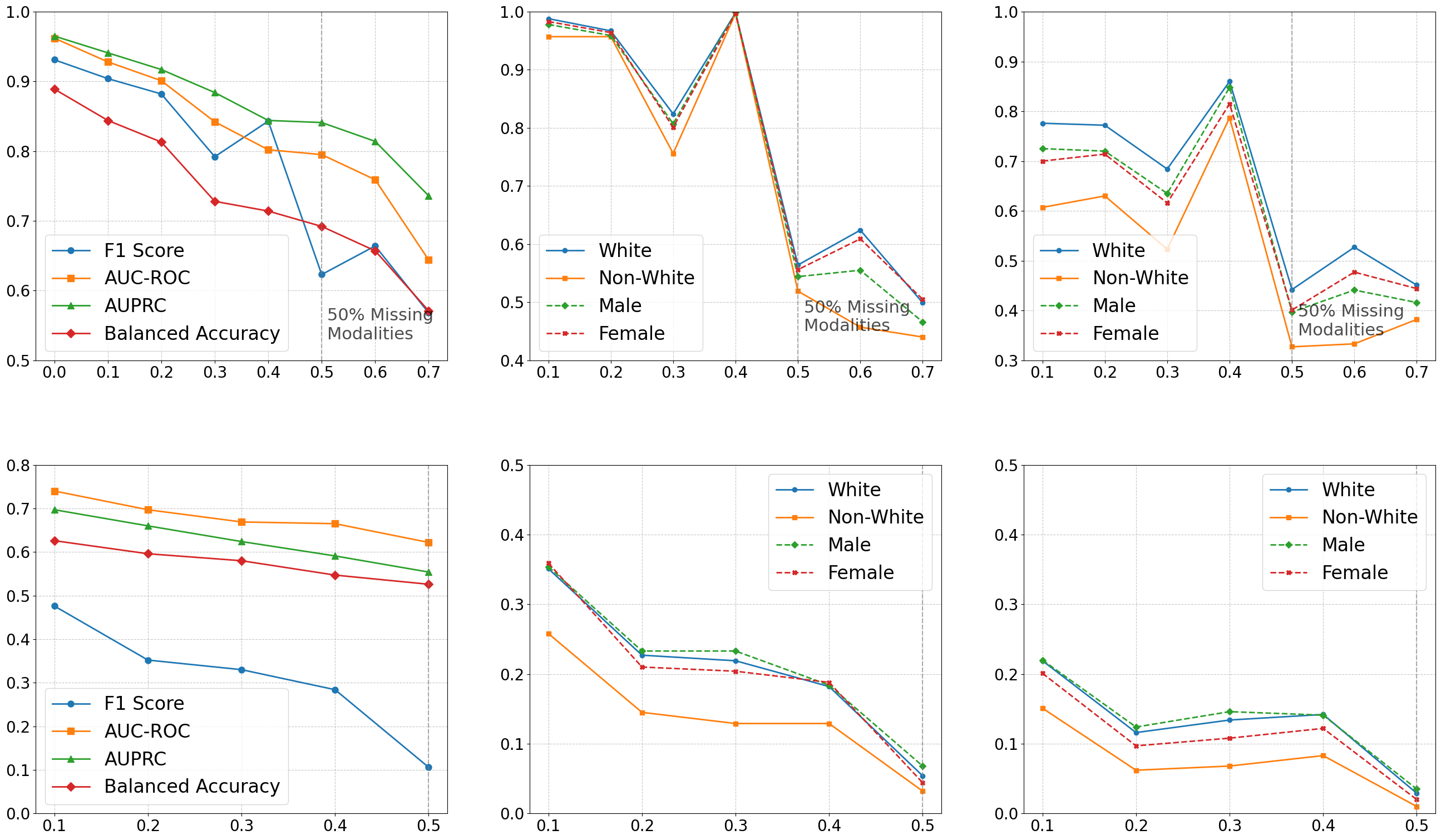} 
    \caption{Model robustness to missing modalities during inference: performance (left), DP (middle), and TPR (right). First row represents results for MIMIC-Eye and second row for MIMIC-IV-Ext-MDS-ED. X-axis represents percentage of missing modalities and  Y-axis represents metrics.  }
    \label{fig:RQ2}
\end{figure}

In RQ2, we evaluate the model's robustness regarding performance and fairness when modalities are absent during inference, as shown in Figure \ref{fig:RQ2}.  Our findings indicate that the absence of modalities during inference negatively impacts the model's overall robustness, leading to a consistent decline in both performance and fairness across DP and TPR metrics. While we do not claim to achieve equal performance when more than 50\% of modalities are missing, it is clear that even a small percentage of missing modalities (0.1\% or 0.2\%), corresponding to one or two modalities rather than the major contributing ones, leads to a noticeable drop in both performance and fairness. This pattern holds true across both datasets.

\vspace{-5pt}

\section{Conclusion}
In conclusion, our findings from RQ1 and RQ2 highlight key considerations for the development and deployment of multimodal models. From RQ1, we observe that adding modalities during training significantly improves model performance, as evidenced by increases in performance measures. However, fairness disparities vary across datasets, emphasizing the need for careful modality selection to balance metrics effectively.

In RQ2, we further identify that even a small percentage of missing modalities during inference consistently leads to a decline in both performance and fairness. This trend raises concerns about the robustness and generalization capability of multimodal models in real-world deployment, where missing modalities are common. We call upon the research community to focus on developing methods to address these challenges, ensuring that future multimodal models are not only performant but also robust and fair in real-world applications, where modality availability can fluctuate.

{
    \small
    \bibliographystyle{ieeenat_fullname}
    \bibliography{main}
}


\end{document}